%% file: NewPaperCameraReady.tex
\documentclass[conference]{IEEEtran}
\IEEEoverridecommandlockouts
\usepackage{cite}
\usepackage{amsmath,amssymb,amsfonts}
\usepackage{algorithmic}
\usepackage{graphicx}
\usepackage{textcomp}
\usepackage{xcolor}
\usepackage{booktabs}
\usepackage{caption}
\usepackage{subcaption}
\usepackage{multirow}
\usepackage{makecell}
\usepackage{float}
\usepackage{diagbox}
\usepackage[accsupp]{axessibility}  

\usepackage[pagebackref,breaklinks,colorlinks]{hyperref}
\usepackage[capitalize]{cleveref}
\crefname{section}{Sec.}{Secs.}
\Crefname{section}{Section}{Sections}
\Crefname{table}{Table}{Tables}
\crefname{table}{Tab.}{Tabs.}

\def\BibTeX{{\rm B\kern-.05em{\sc i\kern-.025em b}\kern-.08em
    T\kern-.1667em\lower.7ex\hbox{E}\kern-.125emX}}
\begin{document}

\title{Masked Collaborative Contrast for Weakly Supervised Semantic Segmentation}

\author{\IEEEauthorblockN{Fangwen Wu}
\IEEEauthorblockA{
\textit{Zhejiang Lab}\\
Hangzhou, China \\
fwu11@zhejianglab.com}
\and
\IEEEauthorblockN{Jingxuan He}
\IEEEauthorblockA{\textit{Zhejiang Lab} \\
Hangzhou, China \\
22021025@zju.edu.cn}
\and
\IEEEauthorblockN{Yufei Yin}
\IEEEauthorblockA{\textit{University of Science and Technology of China} \\
Hefei, China \\
yinyufei@mail.ustc.edu.cn}
\and
\IEEEauthorblockN{Yanbin Hao}
\IEEEauthorblockA{\textit{University of Science and Technology of China} \\
Hefei, China \\
haoyanbin@hotmail.com}
\and
\IEEEauthorblockN{Gang Huang}
\IEEEauthorblockA{\textit{Zhejiang University} \\
Hangzhou, China \\
huanggang@zju.edu.cn}
\and
\IEEEauthorblockN{Lechao Cheng*}$\thanks{*corresponding author.}$
\IEEEauthorblockA{\textit{Zhejiang Lab} \\
Hangzhou, China \\
chenglc@zhejianglab.com}
}

\maketitle

\input{sections/abstract}
\input{sections/introduction}
\input{sections/related_work}

\input{sections/method}

\input{sections/experiment}
\input{sections/limitations}
\input{sections/conclusion}

\section*{Acknowledgement}
 This work is supported by the National Natural Science Foundation of China Grant (No.~U22A6001 and No.~52007173), the Exploratory Research Project of Zhejiang Lab (No.~2022PG0AN01), and the Key Research Project of Zhejiang Lab (No.~2022PG0AC02).

{\small
\bibliographystyle{ieee_fullname}
\bibliography{egbib}
}

\end{document}

%% file: sections/abstract.tex
\begin{abstract}

This study introduces an efficacious approach, Masked Collaborative Contrast (MCC), to highlight semantic regions in weakly supervised semantic segmentation. MCC adroitly draws inspiration from masked image modeling and contrastive learning to devise a novel framework that induces keys to contract toward semantic regions. Unlike prevalent techniques that directly eradicate patch regions in the input image when generating masks, we scrutinize the neighborhood relations of patch tokens by exploring masks considering keys on the affinity matrix. Moreover, we generate positive and negative samples in contrastive learning by utilizing the masked local output and contrasting it with the global output. Elaborate experiments on commonly employed datasets evidences that the proposed MCC mechanism effectively aligns global and local perspectives within the image, attaining impressive performance. The source code is available at \url{https://github.com/fwu11/MCC}.

\end{abstract}

%% file: sections/introduction.tex
\section{Introduction}

Weakly supervised semantic segmentation (WSSS) aims to reduce manual labor in annotating pixel-wise ground-truth labels by using "weak" supervision, such as image-level classification labels~\cite{ahn2018learning,wang2020self,xu2021leveraging,zhang2021weakly, xu2022multi,ru2022learning,chen2023extracting,ru2023token}, points~\cite{bearman2016point, su2022sasformer}, scribbles~\cite{zhang2021dynamic, su2022sasformer} and bounding boxes~\cite{lee2021bbam}. Among all these weak annotations, image-level labels are the most affordable yet challenging, as they only indicate the presence or absence of objects in an image and do not prompt object positions that are fundamental for semantic segmentation. This work focuses on WSSS with only image-level labels.

Previous methods of WSSS with image-level labels typically adopt Class Activation Maps (CAMs)~\cite{zhou2016learning} as pseudo segmentation labels to estimate the locations of target objects approximately. The generated pseudo labels are then refined with diverse approaches~\cite{ahn2018learning,lee2021anti,chen2022class} and further employed to supervise a standard segmentation network. However, the above-mentioned multi-stage framework is usually complicated and suffers from massive burdens, requiring training multiple networks for different subtasks. Another strand of works addresses WSSS in an end-to-end manner~\cite{zhang2020reliability,araslanov2020single,ru2022learning,ru2023token}, \textit{i.e.}, pseudo annotation generation and segmentation prediction are achieved within a single network. Although achieving impressive results, approaches based on single-stage frameworks still confront incomplete object regions since they are shifted by classification procedure where discriminative regions are mainly identified. To alleviate this issue, recent solutions~\cite{ru2022learning,ru2023token} adopt the transformer architecture~\cite{dosovitskiy2020vit} to make full use of the long-range relationships to generate more accurate pseudo labels. However, the crucial relationships between patch tokens have not been well captured semantically~\cite{xue2022protopformer}, which inevitably limits the potential of transformers for WSSS.

Recent studies~\cite{ru2022learning, ru2023token, gao2021ts, xu2022multi, zhu2023weaktr, li2023transcam} have explored the benefits of long-range relationships in Transformer to compensate for discriminative attention (e.g., CAMs). Building upon this, 
we take a further step to ameliorate it with high-level semantics. To achieve this, we first explore the sparsity of the affinity matrix inspired by~\cite{he2023mitigating} and manage to construct meticulous instances that pay much attention to local details within an input.
These local instances are then labeled as positive and negative based on the aggregated activation value of the remaining unmasked tokens. For example, we say the masked instance to be positive if the mean value of remaining tokens in CAMs is larger than the mean value of the whole CAMs. We take advantage of contrastive learning~\cite{he2020momentum} to pull positive instances closer to the global instance and push negative instances away. 

Consequently, representation consistency between global and local instances can be promoted, and semantic discrepancies between foreground and background can be amplified, further facilitating the discovery of accurate and integral objects. In addition, we investigate inter-neighborhood relationships and study different masking strategies for performance, showing a high drop rate (up to 75\%) in the mask with a moderate patch size ($4\times 4$) can better facilitate the performance. 

In summary, our contributions are as follows:
\begin{itemize} 
    \item We introduce Masked Collaborative Contrast (MCC), an effective module to embrace objects of interest by imposing representation consistency between global and local views in the WSSS framework.
    \item We propose an effective single-stage weakly supervised semantic segmentation framework. Unlike existing methods that explicitly erase input image patches to construct local views, we ingeniously integrate the construction of local semantics into Transformer blocks, making it more efficient and allowing salient regions to align better with keys.
    \item Exhaustive experiments showcase that the proposed approach achieves state-of-the-art performance on the prevalent PASCAL VOC \cite{Everingham2010ThePV} and MS COCO \cite{lin2014microsoft} datasets compared with previous single-stage WSSS methods.

\end{itemize}

%% file: sections/related_work.tex
\section{Related Work}
\subsection{Weakly Supervised Semantic Segmentation}

In recent years, a substantial portion of weakly supervised semantic segmentation (WSSS) methods utilizing image-level labels have been developed, with Class Activation Map (CAM) being one of the most commonly used techniques~\cite{zhou2016learning}. Incorporating localization information exclusively from a classification network, CAM activates only the most discriminative regions of an object. To overcome this limitation, various strategies have been proposed, including both multi-stage and end-to-end approaches. These approaches have been proposed to bridge the gap between the classification and the semantic segmentation task by embedding semantics into the learning process, as opposed to only the most discriminative characteristics. 
The standard multi-stage pipeline for WSSS involves first generating seeds via a classification network, then refining the seed to produce a pseudo-mask, and finally training a fully supervised semantic segmentation network with the generated pseudo-mask. Significant attention has been devoted to generating high-quality seeds.
For instance, various approaches~\cite{liexpansion, chen2023extracting, cheng2023out,chen2022class,kho2022exploiting,xu2022multi,su2022re} are introduced to facilitate CAM attention by activating non-discriminative regions or eliminating false positives.
Moreover, pixel-to-image aggregation strategies~\cite{kolesnikov2016seed, jonnarth2022imp, lee2022threshold, lee2021reducing}, additional saliency maps~\cite{lee2021railroad,jiang2022l2g,xu2021leveraging,yao2021non,wu2021embed}, language supervision~\cite{lin2022clip,xie2022clims}, post-processing methods~\cite{ahn2018learning,ahn2019weakly,lee2021anti,chen2022class} have been explored to promote CAM qualities.
Compared with multi-stage methods, single-stage frameworks~\cite{araslanov2020single, ru2022learning,ru2023token,pan2022learning,rossetti2022max} possess the advantage of simplified streamline and advanced efficiency, yet generally sacrifice segmentation accuracy.
1Stage~\cite{araslanov2020single} achieves competitive performance with previous multi-stage methods by designing a segmentation-based network and a self-supervised training scheme.
ViT-PCM~\cite{rossetti2022max} proposes an alternative to CAM based on the locality property of vision transformers through learning a mapping between patch features and classification predictions.
AFA~\cite{ru2022learning} achieves affinity learning during training by imposing constraints on multi-head self-attention.
ToCo~\cite{ru2023token} mines non-discriminative regions by cropping uncertain regions and aligns them with the global object in a contrastive manner. In this work, we draw inspiration from ToCo~\cite{ru2023token} to devise a masked collaborative contrast approach that achieves compelling performance gains. Besides, recent studies~\cite{li2023boosting} also attempt to employ novel unsupervised pre-training solutions for low-data regimes.

\subsection{Transformers in WSSS}
ViT~\cite{dosovitskiy2020vit} revolutionized the vision domain by introducing transformers, resulting in superior performance across a variety of vision tasks. Due to the limited locality of convolutional neural networks, WSSS methods~\cite{ru2022learning, ru2023token, gao2021ts, xu2022multi, zhu2023weaktr, li2023transcam} employ transformer-based approaches to capture global context for its long-range dependency. These methods utilize ViT and its variants as image encoders, leveraging class tokens to predict image labels and generating CAM from patch tokens, delivering impressive results. TS-CAM~\cite{gao2021ts} produces semantic coupled localization maps by coupling class-agnostic attention maps with patch tokens. MCTformer~\cite{xu2022multi} uses multiple class tokens to generate class-specific attention maps and refines CAM with class-specific object localization maps and patch-level pairwise affinity. WeakTr~\cite{zhu2023weaktr}, built upon MCTformer, employs an adaptive attention fusion strategy when combining attention maps from different attention heads. TransCAM~\cite{li2023transcam} refines CAM using attention weights from the transformer while using a dual-branch Conformer~\cite{peng2021conformer} network to generate CAM from the CNN branch, thereby embedding both local features and global representations. This study capitalizes on the global information provided by the self-attention module from the vision transformer and the local information from masked attention, aligning local and global information through contrastive learning.

%% file: sections/method.tex
\section{Method}
\input{sections/fig_framework}
\subsection{Preliminaries} \label{sec:preliminaries}

\noindent \textbf{Masked Self-attention in Transformer.}
Let $I \in \mathbb{R}^{W \times H \times 3}$ be the input image, the vision transformer encoder partitions $I$ into $N = W' \times H'$ non-overlapping patches~$I' \in \mathbb{R}^{N \times (3\times P^2)}$, where $W' = \frac{W}{P}, H' = \frac{H}{P}$, and $P$ is the patch size. The patches $I'$ are then flatted and projected into $N$ patch tokens $\mathbf{x}_0 \in \mathbb{R}^{ N\times D}$, where $D$ represents the dimension of token embedding. The token embeddings, along with the associated learnable class tokens, are concatenated and fed into the standard transformer block. In addition, it is conventional to add an acknowledged effective positional embedding on top of this. Specifically, in the Transformer layer $l$, we first apply linear learnable transformation to map the token sequences to a query matrix $\mathbf{Q}^{(l)} \in \mathbb{R}^{N\times D_k}$, a key matrix $\mathbf{K}^{(l)} \in \mathbb{R}^{N \times D_k}$, and a value matrix $\mathbf{V}^{(l)} \in \mathbb{R}^{N \times D_v}$. Here, $D_k$ denotes the dimensionality of the $\mathbf{Q}$ and $\mathbf{K}$, while $D_v$ is the dimension of $\mathbf{V}$. Consistent with established conventions, we achieve the self-attention mechanism by computing the dot product between each query and all keys, and subsequently scale it by dividing by $\sqrt{D_k}$. This procedure delivers successive affinity matrices $\mathbf{A}^{(l)} \in \mathbb{R}^{N \times N}$ that encompassing pair-wise global relationships in each layer. To regulate these relationships, a binary affinity mask $\mathbf{M}^{(l)}  \in \{0,1\}^{N\times N}$ is introduced as a switch for the affinity matrix, resulting in a regularized affinity matrix $\mathbf{A'}^{(l)} \in \mathbb{R}^{N \times N}$.
The aforementioned association can be expressed in the following manner:
\begin{equation}
\label{eq:mask_attn}
    \mathbf{A'}^{(l)} = \operatorname{softmax} (\mathbf{A}^{(l)} + \mathbf{M'}^{(l)}) \\   
\end{equation} 

\begin{equation}
     \text{where} \ \mathbf{A}^{(l)} = \frac{{\mathbf{Q}^{(l)}}{\mathbf{K}^{(l)}}^\top}{\sqrt{D_k}}
\end{equation}

\begin{equation}
\label{eq:mask}
\mathbf{M'}^{(l)}(x, y) = 
    \begin{cases}
         0 & \textrm{if} \ \ \mathbf{M}^{(l)}(x,y) = 1 \\
         -\infty & \textrm{else}
    \end{cases}
\end{equation}
The output matrix $\mathbf{Z}^{(l)}$ emerges as a weighted amalgamation of $\mathbf{V}^{(l)}$, with the weights being derived from the normalized affinity matrix subjected to a softmax operation. 

\begin{equation}
    \mathbf{Z}^{(l)} = \mathbf{A'}^{(l)} \mathbf{V}^{(l)}
\end{equation}

\noindent \textbf{CAM Generation.}
We employ Class Activation Maps (CAMs)~\cite{zhou2016learning} to derive initial pseudo segmentation labels. Assuming that the output sequence of patch tokens at each transformer layer is reshaped to a feature map $\mathbf{Z}^{(l)} \in \mathbb{R}^{D\times H' \times W'}$, where $D$ is the feature dimension, and $H' \times W'$ is the spatial dimension. CAMs at the $l$-th layer $\mathbf{F}^{(l)} \in \mathbb{R}^{C \times H' \times W'}$ are then generated through matrix multiplication between the feature map and the parameters $\mathbf{W}^{(l)} \in \mathbb{R}^{D \times C}$ of the corresponding classifier:
\begin{equation}
    \mathbf{F}^{(l)} = {\mathbf{W}^{(l)}}^{\top} \mathbf{Z}^{(l)}
\end{equation}
where $C$ denotes the number of foreground categories.

\input{sections/fig_mask}
\subsection{Overview}
Figure \ref{fig:framework} demonstrates the overall single-stage framework for WSSS. Following the common practice~\cite{ru2022learning, ru2023token}, we adopt a standard vision transformer as the encoder to realize classification and a lightweight decoder to make segmentation predictions. For classification, representations of patch tokens from the last layer are aggregated through Global Maximum Pooling (GMP) followed by a convolutional classifier, producing the class score vector $\hat{y}$ and the multi-label soft margin loss is applied to calculate the classification loss $\mathcal{L}_{\text{cls}}$. 
\begin{multline} \label{eq:cls}
    \mathcal{L}_{\text{cls}} = -\frac{1}{C}\sum_{c=1}^Cy^c \log(\frac{1}{1+ \exp{(-\hat{y}^c)}}) \\
    +(1-y^c)\log(\frac{\exp{(-\hat{y}^c)}}{1+ \exp{(-\hat{y}^c})})
\end{multline}
where $C$ is the number of classes and image-level ground-truth labels $y$ .
The CAM derived from the classifier is refined with PAR proposed in~\cite{ru2022learning} to a reliable pseudo segmentation label, which is subsequently used to supervise segmentation predictions, producing the segmentation loss $\mathcal{L}_{\text{seg}}$. 
We also leverage patch tokens from a predefined intermediate layer to generate an auxiliary CAM. These tokens are further processed to pseudo affinities as guidance for pairwise relations of final patch tokens to alleviate over-smoothing, as suggested in~\cite{ru2023token}. The process yields the auxiliary classification loss $\mathcal{L}_{\text{cls}}^{\text{aux}}$, calculated using the multi-label soft margin loss, as well as the affinity loss $\mathcal{L}_{\text{aff}}$. In addition, various binary masks are applied to self-attention in the transformer encoder to generate a number of class tokens that only contain local information. These local class tokens are assigned with positive/negative in virtue of activations of the auxiliary CAM. Representation consistency of global and local class tokens are optimized by the contrastive loss $\mathcal{L}_{\text{mcc}}$. In summary, the overall training objective is:
\begin{equation} \label{eq:3}
    \mathcal{L} = \underbrace{\mathcal{L}_{\text{cls}} + \mathcal{L}_{\text{cls}}^{\text{aux}}+ \mathcal{L}_{\text{seg}}}_{\mbox{baseline}}
 + \lambda_a\mathcal{L}_{\text{aff}}+\lambda_m\mathcal{L}_{\text{mcc}} 
\end{equation}
where $\lambda_a$ and $\lambda_m$ are the weighting factors for losses.

\subsection{Masked Collaborative Contrast}
Although the quality of CAMs is ameliorated by virtue of the long-range modeling capabilities of transformers, there are still some non-discriminative object regions that need to be identified to elevate segmentation performance. Inspired by Masked Image Modelling~\cite{xie2022simmim} and Contrastive Learning~\cite{oord2018representation,he2020momentum, caron2021emerging}, we design a novel module, termed Masked Collaborative Contrast (MCC), to achieve more integral object coverage by imposing representation consistency between global and local views of the same input image.

The proposed module employs binary masks to extract local information, which entails the selective elimination of columns within the affinity matrix during the attention operation. This process effectively filters out specific keys, compelling the transformer encoder to focus its attention on the remaining tokens. A noteworthy aspect of this masking procedure is its application at a reduced resolution through downsampling. Subsequently, the final target mask matrix is synthesized by upsampling to form contiguous, squared, and localized masking regions. The masks generated during the attention operation serve the crucial role of distinguishing between positive and negative local class tokens that encapsulate high-level semantic information. MCC aims to optimize the similarity between the global class token and the positive/negative local class tokens within the latent space, and this optimization is achieved through the utilization of a contrastive loss. This comprehensive approach not only enhances the model's ability to capture essential local information but also improves the discrimination capabilities in the latent space, thereby contributing to the overall performance and robustness of the model. We will clarify the details in the following paragraph.

\noindent \textbf{Random Masking with Keys.} MCC employs binary masks to extract local information through successive affinity matrices. As outlined in Equations~\ref{eq:mask_attn} to \ref{eq:mask}, a binary affinity mask $\mathbf{M} \in \{0,1\}^{N\times N}$ is introduced to regulate the information flow within the affinity matrix $\mathbf{A} \in \mathbb{R}^{N \times N}$. For simplicity, we omit the layer index in the discussion. The intricacies of our proposed masking strategy are depicted in Figure~\ref{fig:mask}. Specifically, masking is achieved by selectively "dropping" columns within the affinity matrix. This process essentially involves discarding specific keys, compelling the transformer encoder to prioritize the analysis of the remaining tokens. It is essential to note that the keys designated for masking are independently sampled from a Bernoulli distribution with a masking ratio $p \in (0,1)$ applied along the token dimension. Subsequently, these masked keys are expanded to match the dimensions of the affinity matrix, thus culminating in the creation of $\mathbf{M} \in \{0,1\}^{N \times N}$.

\noindent \textbf{Random Masking with Scales.}
Figure~\ref{fig:mask} (right) illustrates the versatility of the proposed random masking technique, showcasing its applicability across varying scales. This process entails two main steps: downsampling, which adheres to a predetermined multiple of the specified scale, is employed to generate the initial mask matrix at the lower resolution; subsequently, upsampling the initial mask matrix by the same preset multiple results in the formation of the final target mask matrix. A salient aspect to observe is that, due to the upsampling procedure, the elements within the initial mask matrix coalesce to form contiguous, squared, and localized masks.

\noindent \textbf{Positive/negative Determination.}
We define positive and negative local images based on activation values of the auxiliary CAM and attention masks. Concretely, let $\mathbf{M}_{t} = \Gamma^{N \rightarrow H' \times W'} (\mathbf{M}_{i:})$ be a key mask derived from the corresponding affinity mask, where $i$ is an arbitrary row index and $\Gamma(\cdot)$ denotes the reshape operator. Here we use $1$ to represent a masked token and $0$ to be an unmasked token. Intuitively, the remaining tokens are likely to belong to semantic objects if their averaged activation values are high. As a consequence, we distinguish positiveness from negativeness by the following equation:
\begin{multline}
\small
\label{eq:pos}
       \text{positiveness(+)} := \\ 
       \mathbb{I} (\frac{1}{N}\sum{\mathbf{Y'}_t \odot (1 - \mathbf{M}_t}) > \mu \frac{1}{N} \sum{(1 - \mathbf{M}_t)}) 
\end{multline}

\noindent where $\mathbb{I}(\cdot)$ is an indicator function, $\odot$ denotes Hadamard product, $\mu$ is a predefined threshold for positiveness and $\mathbf{Y'}_t$ is a discretized token-level label defined as:

\begin{equation} 
\label{eq:discrete}
    \mathbf{Y'}_t = 
    \begin{cases}
    2 & \text{if} \max_c(\mathbf{F}_{\text{aux}}) \geq \beta_{\text{fg}}\\ 
    0 & \text{if} \max_c(\mathbf{F}_{\text{aux}}) \leq \beta_{\text{bg}}\\
    1 & \text{else}\\
    \end{cases}
\end{equation}

\noindent where  from activation values given two thresholds $\beta_{\text{bg}}$ and $\beta_{\text{fg}}$ and $\mathbf{F}_{\text{aux}}$ denotes the feature map derived from the auxiliary layer to ameliorate the issue of over-smoothing.

\input{sections/fig_fg_bg}

\noindent \textbf{Contrastive Loss.}
We employ InfoNCE loss~\cite{oord2018representation} for contrastive learning. Global and local class tokens are linearly projected to a latent space appropriate for contrasting through global and local projectors, respectively. Let $q$ be the projected global class token and $k^+/k^-$ be the projected positive/negative local class tokens. The training objective is to optimize the similarity between the global class token and positive/negative local class tokens:
\begin{multline} \label{eq:8}
\small
    \mathcal{L}_{mcc} = \\
    -\frac{1}{N^{+}} \sum_{k^+} \log\frac{\exp(qk^+/\tau)}{\exp(qk^+/\tau) + \sum_{k^-} \exp(qk^-/\tau) + \epsilon}
\end{multline}
where $N^+$ counts the number of $k^+$ samples, $\tau$ is the temperature factor, and $\epsilon$ is introduced for numerical stability. Parameters of the global projector are updated using the moving average strategy proposed in MoCo~\cite{he2020momentum}, \textit{i.e.}, $\theta_g \leftarrow m \theta_g + (1-m)\theta_l$, where $m$ is the momentum factor, $\theta_g$ and $\theta_l$ are the parameters of the global projector and the local projector, respectively. This slowly evolved update on parameters of two projectors ensures training stability and enforces representation consistency.

\subsection{Affinity Learning}
In this section, we introduce affinity learning to facilitate pairwise relations of final patch tokens, as suggested in~\cite{ru2023token}. It is observed that patch tokens from deeper transformer layers suffer from over-smoothing~\cite{shi2022revisiting, he2023mitigating}, and the resultant unified representations impair semantic segmentation performance severely. To address this issue, we first derive favorable CAMs from an intermediate layer, then leverage them as available constraints to optimize pairwise relations of final patch tokens.


\noindent \textbf{Affinity Label Generation.}
We first aggregate patch tokens from a chosen intermediate layer through global max-pooling (GMP), as suggested in~\cite{ru2022learning}, then apply a convolution layer to generate classification logits. The intermediate CAMs $\mathbf{F}_{\text{aux}}$ are subsequently processed to pseudo affinity labels as supervision for affinity learning. Note that it is difficult to accurately differentiate foreground from background based on the activation values of the derived CAMs, since pixels with medium confidence are inappropriate to be labeled as either an annotated object or background.
To generate a reliable affinity label, we introduce two thresholds $\beta_{\text{fg}}$ and $\beta_{\text{bg}}$ satisfying $0 < \beta_{\text{bg}} < \beta_{\text{fg}} < 1$ to partition CAMs into foreground, background and uncertain regions. Mathematically, the reliable segmentation label $\mathbf{Y'} \in \mathbb{R}^{H' \times W'}$ is formed as follows:
\begin{equation} \label{eq:thr}
    \mathbf{Y'} = 
    \begin{cases}
    \operatorname*{argmax}_{c} (\mathbf{F}_{\text{aux}}) & \text{if} \max_c(\mathbf{F}_{\text{aux}}) \geq \beta_{\text{fg}}\\ 
    0 & \text{if} \max_c(\mathbf{F}_{\text{aux}}) \leq \beta_{\text{bg}}\\
    255 & \text{else}\\
    \end{cases}
\end{equation}
Here we label the background class as 0 and the uncertain region as 255. Pairwise affinity relations are then constructed based on this pseudo segmentation label. Concretely, we determine the affinity to be positive if the pixel pairs sampled from the segmentation label share the same semantic (\textit{e.g.}, the pixel pairs are both from foregrounds or both from backgrounds); otherwise, their affinity is treated as negative. Affinities will be ignored when pixels are sampled from uncertain regions.

\noindent \textbf{Affinity Loss.}
The reliable affinity relations are then harnessed as supervision to promote representations of patch tokens from the last layer of the transformer encoder. In addition, we use the cosine similarity to measure the predicted affinity between two final patch tokens.
The affinity loss is therefore calculated as:
\begin{multline}
\label{eq:6}
    \mathcal{L}_{aff} = \frac{1}{N^+}\sum_{\mathbf{Y}_i = \mathbf{Y}_j}(1-\cos(\mathbf{T}^{(L)}_{:,i},\mathbf{T}^{(L)}_{:,j})) \\
    + \frac{1}{N^-}\sum_{\mathbf{Y}_i \neq \mathbf{Y}_j}\cos(\mathbf{T}^{(L)}_{:,i},\mathbf{T}^{(L)}_{:,j})
\end{multline}

\noindent where $\mathbf{T}^{(L)} = \Gamma^{D\times H' \times W' \rightarrow D\times N}(\mathbf{Z}^{(L)})$, $\mathbf{Y} = \Gamma^{H' \times W' \rightarrow N}(\mathbf{Y'})$. $\Gamma(\cdot)$ is the reshape operator, $\cos(\cdot, \cdot)$ denotes the cosine function, and $N^+$/$N^-$ count the number of positive/negative samples.
Intuitively, this objective function directly encourages final patch tokens with a positive relation to be more similar, and otherwise be more distinctive. It also benefits the learning of token representations from the earlier transformer layers according to the chain rule.

%% file: sections/fig_framework.tex
\begin{figure*}[!ht]
\centering
\includegraphics[trim={0.5cm 0 2cm 0},clip,width=0.90\textwidth]{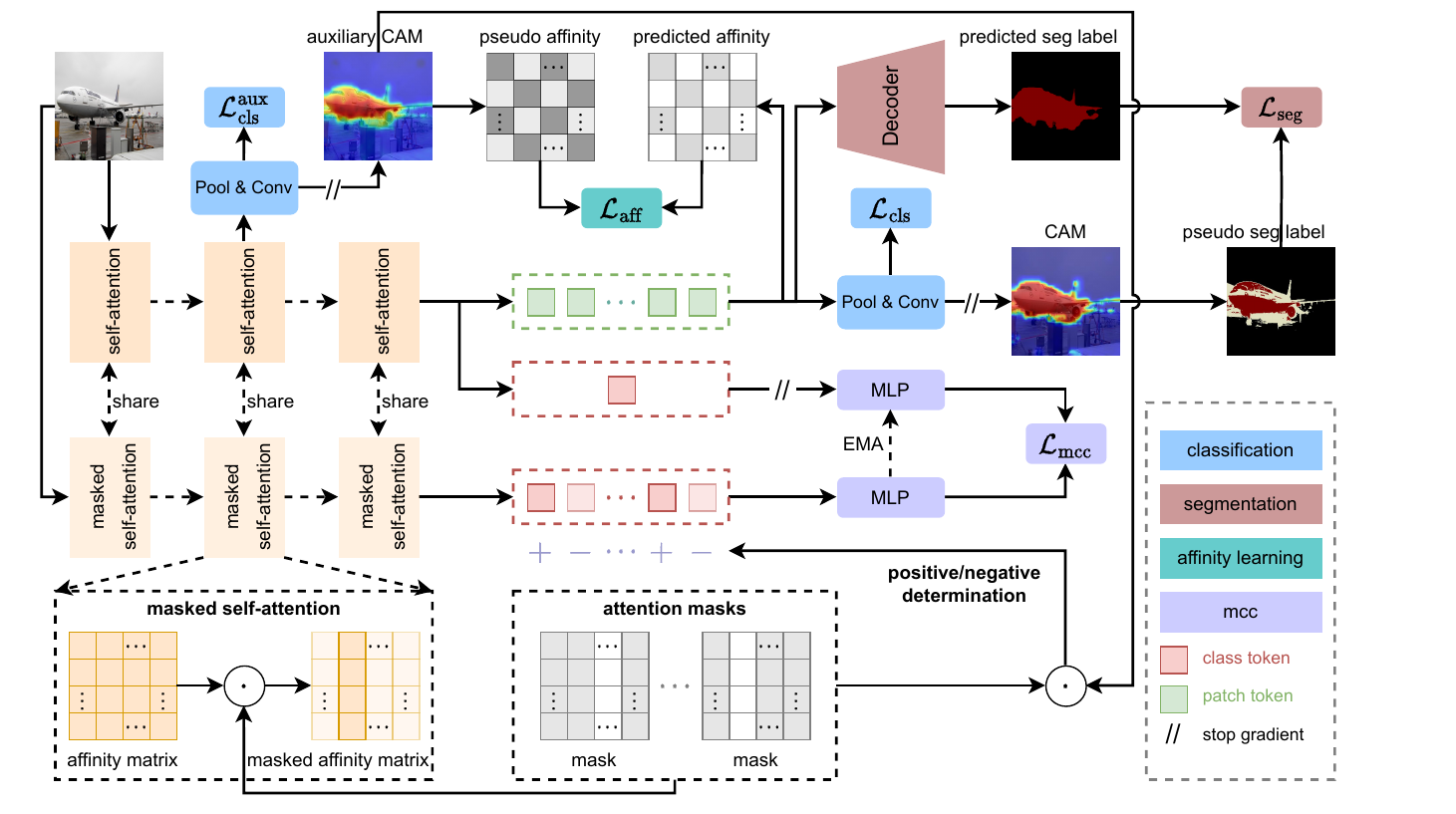}
\caption{\small Visualization of our single-stage framework for WSSS. It is composed of classification, segmentation, affinity learning, and masked collaborative contrast. Particularly, we generate a random mask and integrate it into the transformer encoder to produce a local class token. Several local class tokens are obtained in the same way. After positive/negative determination, they are aligned with the global class token by contrastive learning.}
\label{fig:framework}
\vspace{-5mm}
\end{figure*}

%% file: sections/fig_mask.tex
\begin{figure*}
\centering
\begin{subfigure}{0.49\textwidth}
\centering
\includegraphics[height=2.0in, width=3.2in]{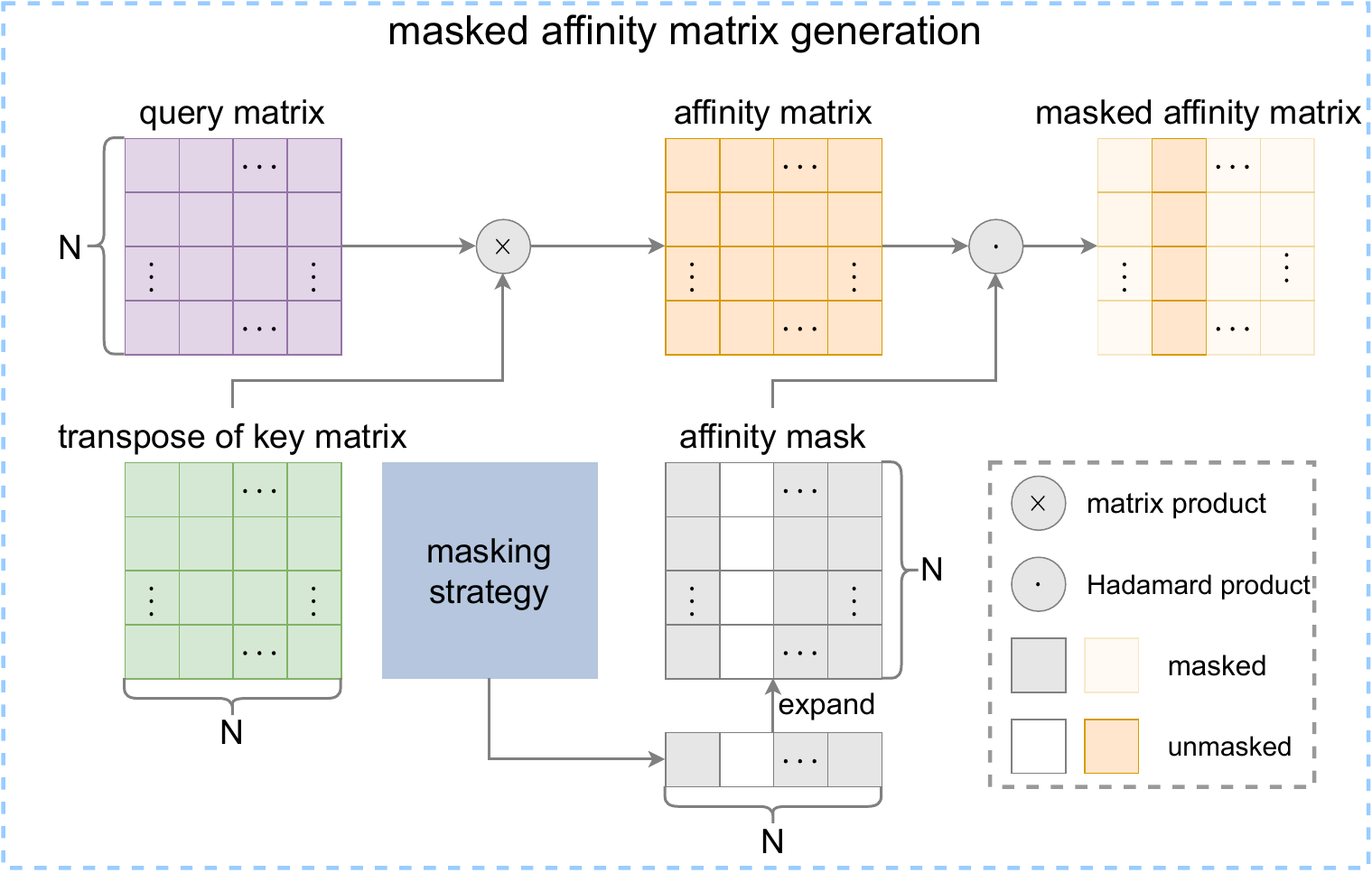}
\end{subfigure}
\begin{subfigure}{0.49\textwidth}
\centering
\includegraphics[height=2.0in, width=3.2in]{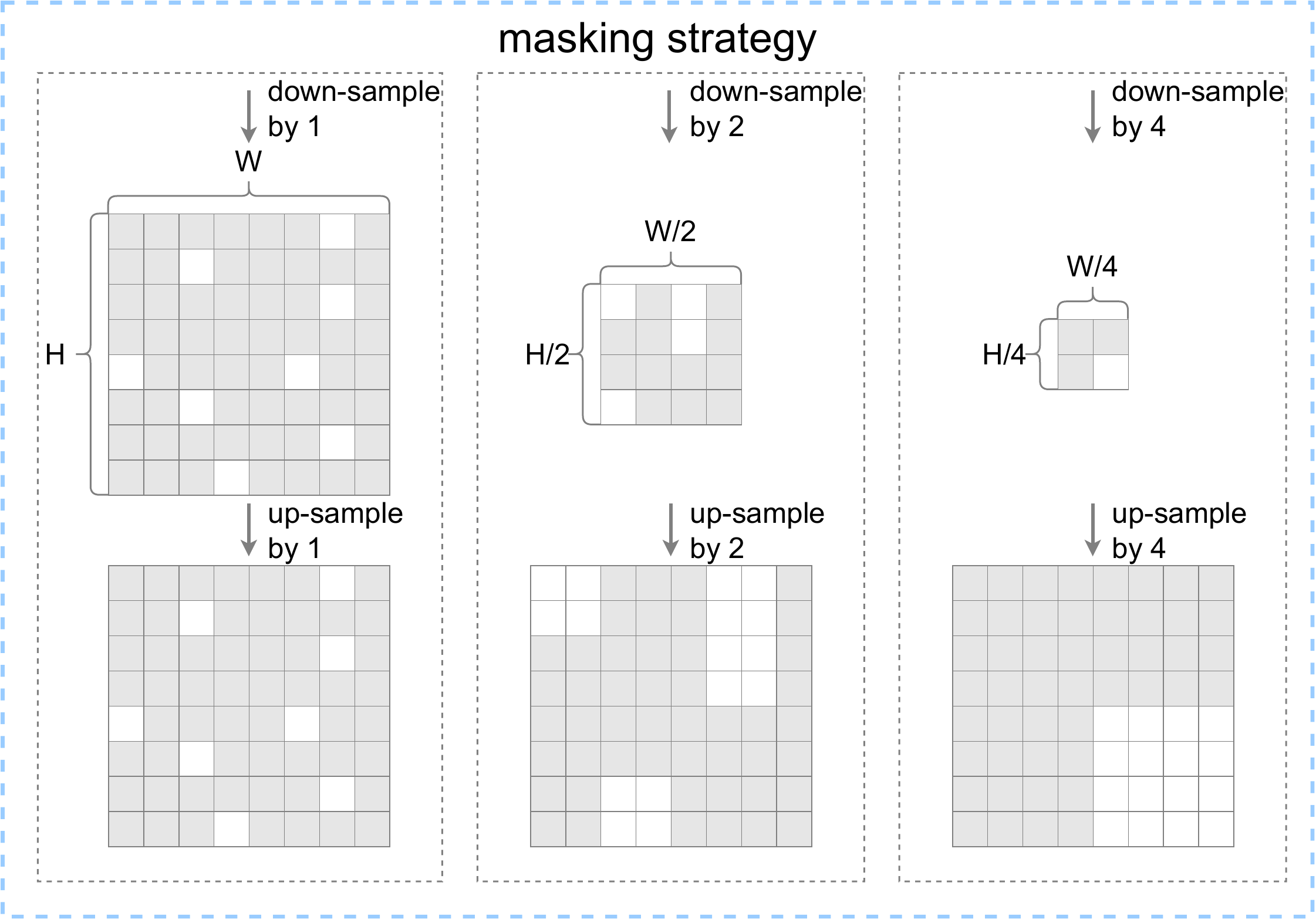}
\end{subfigure}
\caption{\small Visualization of the generation process of masked affinity matrix (left) and the masking strategy (right).}
 \vspace{-5mm}
\label{fig:mask}
\end{figure*}

%% file: sections/fig_fg_bg.tex
\begin{figure}[!ht]
\centering
\includegraphics[width=0.45\textwidth]{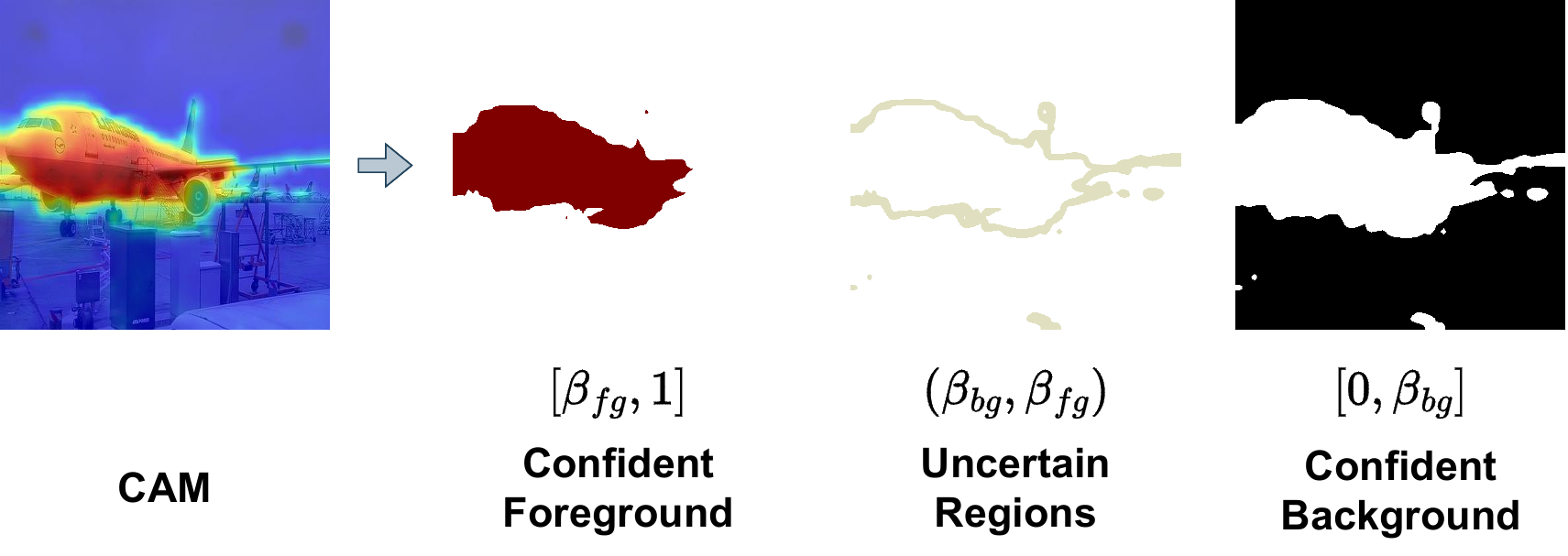}
\caption{\small The generated CAM is segregated into three distinct segments, utilizing thresholds defined by $\theta_{bg}$ and $\theta_{fg}$. This partitioning process enables the accurate identification of positive and negative labels, as well as the subsequent generation of affinity labels.}
\label{fig:fg_bg}
\vspace{-3mm}
\end{figure}

%% file: sections/experiment.tex
\section{Experiments}
\subsection{Setup}
\noindent \textbf{Datasets and Metrics.}
We evaluate our proposed method on two commonly used benchmark datasets, \textit{i.e.}, PASCAL VOC~\cite{Everingham2010ThePV} and MS COCO~\cite{lin2014microsoft}, which contains 20 and 80 distinct object classes, respectively. For PASCAL VOC, we adopt the prevailing SBD dataset~\cite{6126343} with 10,582 training images in our experiments, following the common practice \cite{chen2022class, xu2022multi}. It also has 1,449 for validation and 1,456 images for testing.  MS COCO has 82,081 training images and 40,137 validation images. 
For all experiments, we only use image-level labels for training. 
We apply mean Intersection over Union (mIoU) as the evaluation metric.

\input{sections/tab_voc_coco}

\noindent \textbf{Implementation Details.}
We use DeiT-base \cite{touvron2021training} pre-trained on ImageNet-1K \cite{russakovsky2015imagenet} as our transformer backbone. 
For PASCAL VOC, the network is trained with a batch size of 4 for a total of $20k$ iterations. For MS COCO, the network is trained for $80k$ iterations with a batch size of 8. 
AdamW \cite{loshchilov2018decoupled} is used for optimization with a polynomial scheduler. The initial 1500 iterations are considered as a warm-up stage, during which the learning rate is set to $1e^{-6}$, and gradually increased in a linear fashion to reach $6e^{-5}$. In the following iterations, the learning rate decays with a rate of $0.9$.  Training images are cropped into $448 \times 448$. In affinity learning procedure, the auxiliary layer is chosen to be the 10th layer of the Transformer block as suggested in ToCo \cite{ru2023token}. $\beta_{\text{bg}}$ and $\beta_{\text{fg}}$ in Equation \ref{eq:discrete} and \ref{eq:thr} are set as 0.25 and 0.7.
In MCC module, the masking ratio and the masking scale are 0.75 and 4, respectively. The threshold $\mu$ for positiveness in Equation \ref{eq:pos} is 0.2. The temperature factors $\tau$ in Equation \ref{eq:8} is set as 0.5. The momentum factor $m$ for the EMA process is chosen to be 0.9. The loss weights $\lambda_a$ and $\lambda_m$ in Equation \ref{eq:3} are set as 0.2 and 0.5, respectively. During inference, we use multi-scale testing and dense-CRF processing, as suggested in \cite{CP2015Semantic}.

\subsection{Comparison with State-of-the-arts}
\input{sections/tab_pesudo}
\input{sections/fig_cam_results}

\noindent\textbf{Pseudo Labels Performance.} Table \ref{table:pesudo} presents the quantitative pseudo label results on PASCAL VOC dataset in terms of mIoU. Notably, our method can produce better pseudo labels than other Transformer-based single-stage WSSS methods like AFA \cite{ru2022learning} and ToCo \cite{ru2023token} with ImageNet-1k pretrained weights. It is worth noting that our performance even surpasses ViT-PCM \cite{rossetti2022max} which uses ImageNet-21k pretrained weights.

The qualitative results on both PASCAL VOC and MS COCO datasets are presented in the Figure \ref{fig:cam_results} and Figure \ref{fig:coco_results_1}, respectively. These results  demonstrate that the CAMs generated with our method can cover more integral part of the objects while reducing the noisy components, compared to other Transformer-based end-to-end SOTA WSSS methods.
\input{sections/fig_coco_results}

\noindent\textbf{Segmentation Performance.}
Table \ref{table:voc-coco} presents the quantitative semantic segmentation results on the PASCAL VOC and MS COCO datasets. Our proposed method outperforms existing competitive single-stage WSSS methods, such as SLRNet \cite{pan2022learning} and ToCo~\cite{ru2023token}, on both the PASCAL VOC and MS COCO datasets. It shows remarkable performance on online PASCAL VOC testing, achieving a score of 71.2\% compared to 70.5\% for ToCo, and delivers an improvement of up to 1.0\% compared to ToCo on the large-scale MS COCO dataset. Furthermore, our proposed method is on par with many multi-stage WSSS methods that employ additional supervision, such as saliency maps and natural language, demonstrating its sufficiency during the training phase.


\input{sections/fig_seg_results}
Figure~\ref{fig:seg_results} and Figure \ref{fig:coco_results_2} shows the visualization of the generated segmentation results with our proposed MCC in PASCAL VOC and MS COCO dataset respectively. Compared with previous methods, our model succeeds in segmenting multiple objects in an image with more complete and accurate object boundaries, even when the objects are small. Furthermore, our model is able to distinguish objects from background areas with similar colors.

\subsection{Ablation Study}

\noindent\textbf{Improvements of Masked Collaborative Contrast.} We provide quantitative results of CAM and Segmentation on the PASCAL VOC \emph{train} and \emph{val} set in Table \ref{table:ablation}. Our baseline model incorporates a classifier subsequent to the transformer encoder, an auxiliary classifier, derived from an intermediate layer, in conjunction with a decoder designated for segmentation. The introduction of a token affinity learning module, addressing the over-smoothing problem, results in a marked enhancement in performance relative to the baseline model. The incorporation of our proposed MCC module yields considerable enhancements in the segmentation performance, with gains exceeding 4.0\% within the training set and 2.0\% within the validation set. After applying CRF post-processing, the performance finally boosts to the 70.3\% mIoU on the \emph{val} set.

\input{sections/tab_ablation}

The proposed MCC module attempts to attend objects of interest by imposing representation consistency between global and local views based on masked attention, which differs from existing approaches that explicitly erase image patches as the input, making it more efficient and allowing salient regions to align better with keys.
\input{sections/fig_scale_ratio}

\noindent \textbf{Mask Sampling Strategy.} During the process of random masking, the masking ratio and masking scale play pivotal roles in controlling the extent and resolution of masked patches, respectively. As shown in Figure~\ref{fig:abl_mask}, the sensitivity analysis sheds light on the ramifications of varying configurations for the masking scale and masking ratio. The left segment of Figure~\ref{fig:abl_mask} delineates that, among the entire spectrum of masking ratios spanning from 0.45 to 0.99, a masking scale of 4 emerges as the optimal choice for average. On the opposite side, the right segment of Figure~\ref{fig:abl_mask} signifies that, within the range of masking scales (1, 2, 4, and 7), a masking ratio of 0.75 holds sway as the most advantageous for averaged scenarios. The judicious selection of values for both the masking scale and masking ratio orchestrates preserving long-range connections while concurrently upholding the integrity of local patches. This empowers the network to extract sophisticated features from a localized vantage point, thereby amplifying its comprehensive performance. 

\input{sections/tab_fg_bg_thres}
\noindent \textbf{Foreground/Background Thresholds.} In Table \ref{table:fg_bg_thres}, we report the impact of using different background thresholds to differentiate between foreground, background, and uncertain regions. Our results show that the combination of $\beta_{fg} = 0.7$ and $\beta_{bg} = 0.25$ achieves the best performance.

%% file: sections/tab_voc_coco.tex
\begin{table}[htbp]
\caption{\small Segmentation results in terms of mIoU($\%$) on PASCAL VOC and MS COCO datasets. $\mathcal{I}$, $\mathcal{S}$, and $\mathcal{L}$ denote image-level labels, the external saliency maps and external language supervision used for supervision, respectively.}
\renewcommand\arraystretch{1.2}
\begin{center}
\resizebox{1.0\linewidth}{!}{
\begin{tabular}{l |c|c|c c| p{1cm}<{\centering}}
\hline
Methods & Backbone & Sup. & \multicolumn{2}{c|}{VOC} & COCO  \\
 &  &  &  \emph{val} &  \emph{test} &  \emph{val} \\

\hline
\multicolumn{5}{l}{\textbf{Multi-Stage WSSS Methods}} \\
\hline 
AuxSegNet \cite{xu2021leveraging} & WResNet38 & \multirow{3}*{$\mathcal{I,S}$} & 69.0 & 68.6 & 33.9\\
EPS \cite{lee2021railroad} & ResNet101 & ~ & 71.0 & 71.8 & 35.7 \\
L2G \cite{jiang2022l2g}& ResNet101 & ~ & 72.1 & 71.7 & 44.2 \\
\hline
CLIMS \cite{xie2022clims} & ResNet101 & \multirow{2}*{$\mathcal{I,L}$} & 69.3 & 68.7 & - \\
CLIP-ES \cite{lin2022clip} & ResNet101 &  ~  & 71.1& 71.4 & 45.4\\
\hline
ReCAM \cite{chen2022class} & ResNet50 & \multirow{5}*{$\mathcal{I}$} & 68.5 & 68.4 & 45.0\\
AMR \cite{Qin_Wu_Xiao_Li_Wang_2022} & ResNet50 & ~ & 68.8 & 69.1 & -\\
ESOL \cite{liexpansion} & ResNet50 & ~ & 69.9 & 69.3 & 42.6\\
AMN \cite{lee2022threshold} & ResNet50 & ~ & 69.5 & 69.6 & 44.7\\
MCTformer \cite{xu2022multi} & DeiT-S & ~ & 71.9 & 71.6 & 42.0\\
\hline
\multicolumn{5}{l}{\textbf{Single-Stage WSSS Methods}} \\
\hline
1Stage \cite{araslanov2020single} & WResNet38 & \multirow{5}*{$\mathcal{I}$} & 62.7 & 64.3 & - \\
AFA \cite{ru2022learning} & MiT-B1 & ~ & 66.0 & 66.3 & 38.9\\
SLRNet \cite{pan2022learning} & WResNet38 & ~ & 67.2 & 67.6 & 35.0\\
ToCo \cite{ru2023token} & DeiT-B & ~ & 69.8 & 70.5 & 41.3\\
MCC (Ours) & DeiT-B & ~ & \textbf{70.3} & \textbf{71.2} & \textbf{42.3}\\
\hline
\end{tabular}
}
 \vspace{-3mm}
\end{center}
\label{table:voc-coco}
\end{table}

%% file: sections/tab_pesudo.tex
\begin{table}
\caption{\small Pseudo labels evaluation results compared to other single-stage WSSS methods in terms of mIoU($\%$) on the PASCAL VOC dataset.}
\renewcommand\arraystretch{1.2}
\begin{center}
\resizebox{1.0\linewidth}{!}{
\begin{tabular}{l|c|c c}
\hline
Methods & Backbone & VOC \emph{train} & VOC \emph{val} \\
\hline
1Stage \cite{araslanov2020single}$_{\text{CVPR'2020}} $& WResNet38 & 66.9 & 65.3 \\
SLRNet \cite{pan2022learning}$_{\text{IJCV'2022}}$ & WResNet38 & 67.1 & 66.2 \\
AFA \cite{ru2022learning}$_{\text{CVPR'2022}}$ & MiT-B1 & 68.7 & 66.5 \\
ViT-PCM\cite{rossetti2022max}$_{\text{ECCV'2022}} $ & ViT-B & 71.4 & 69.3 \\
ToCo \cite{ru2023token}$_{\text{CVPR'2023}}$ & Deit-B & 72.2 & 70.5 \\
MCC(Ours) & Deit-B & \textbf{73.0} & \textbf{71.3} \\
\hline
\end{tabular}
}
 \vspace{-5mm}
\end{center}
\label{table:pesudo}
\end{table}

%% file: sections/fig_cam_results.tex
\begin{figure*}[h]
\centering
\includegraphics[trim={0.5cm 0 0 0.5cm},clip,width=0.9\linewidth,height=0.4\linewidth]{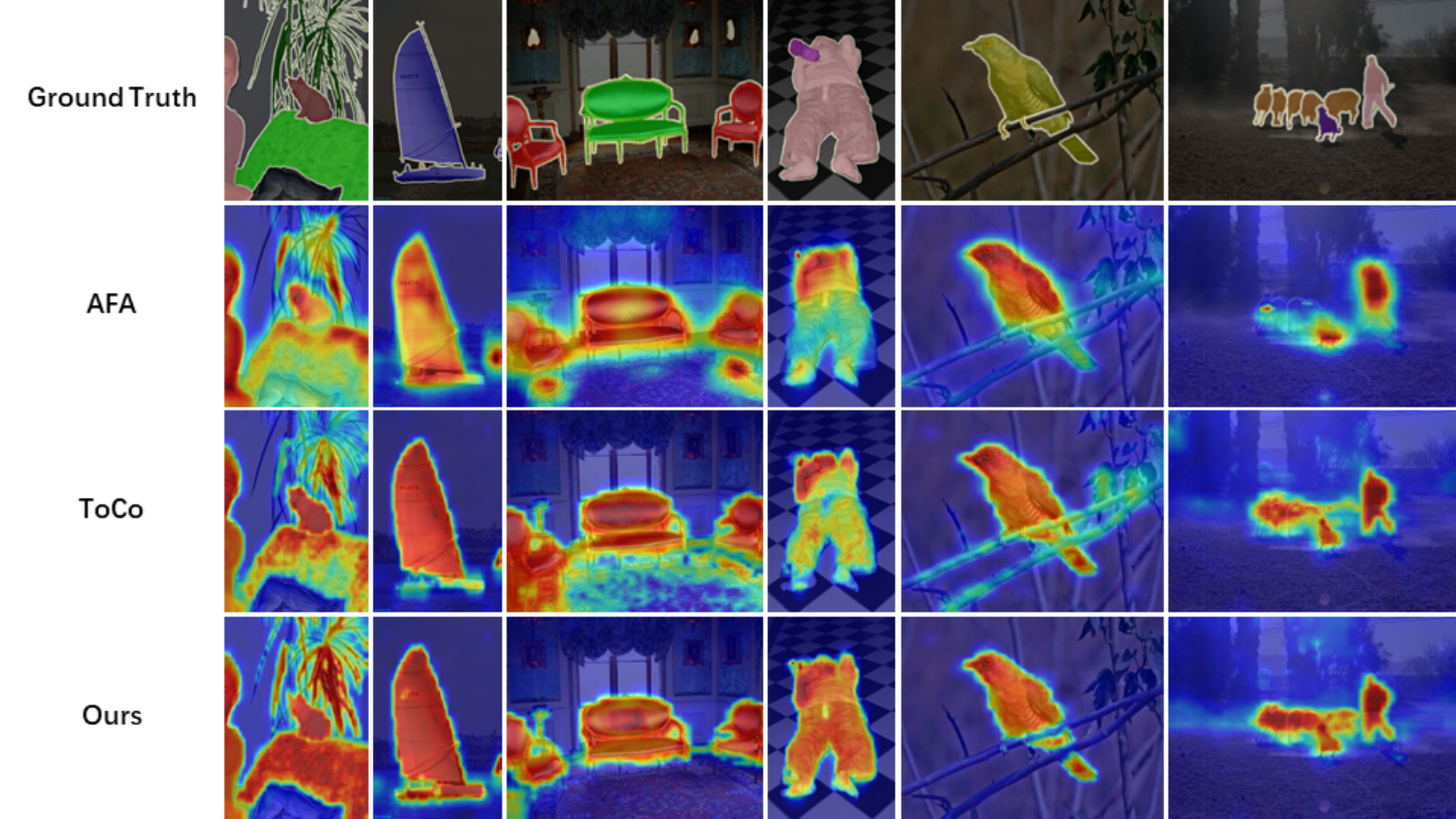}
\caption{\small CAM results of Transformer-based end-to-end WSSS methods AFA \cite{ru2022learning}, ToCo~\cite{ru2023token} and our method on PASCAL VOC \emph{val} set.}
\label{fig:cam_results}
\end{figure*}

%% file: sections/fig_coco_results.tex
\begin{figure}[!htb]
\centering
\begin{subfigure}{0.45\textwidth}
\centering
\includegraphics[height=1.2in,width=3.0in]{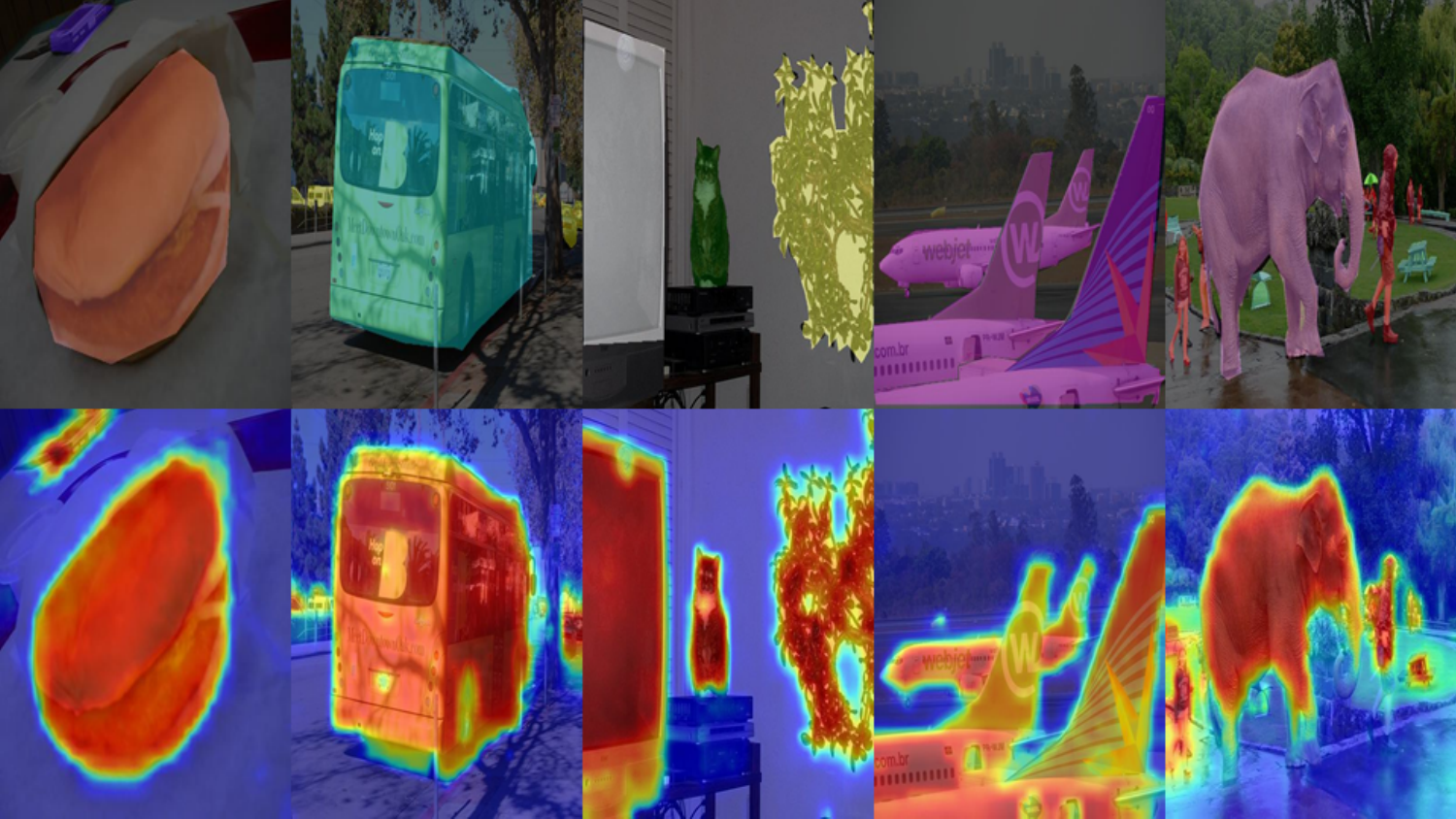}
\caption{\small CAM results. }
\label{fig:coco_results_1}
\end{subfigure}
\begin{subfigure}{0.45\textwidth}
\centering
\includegraphics[height=1.2in, width=3.0in]{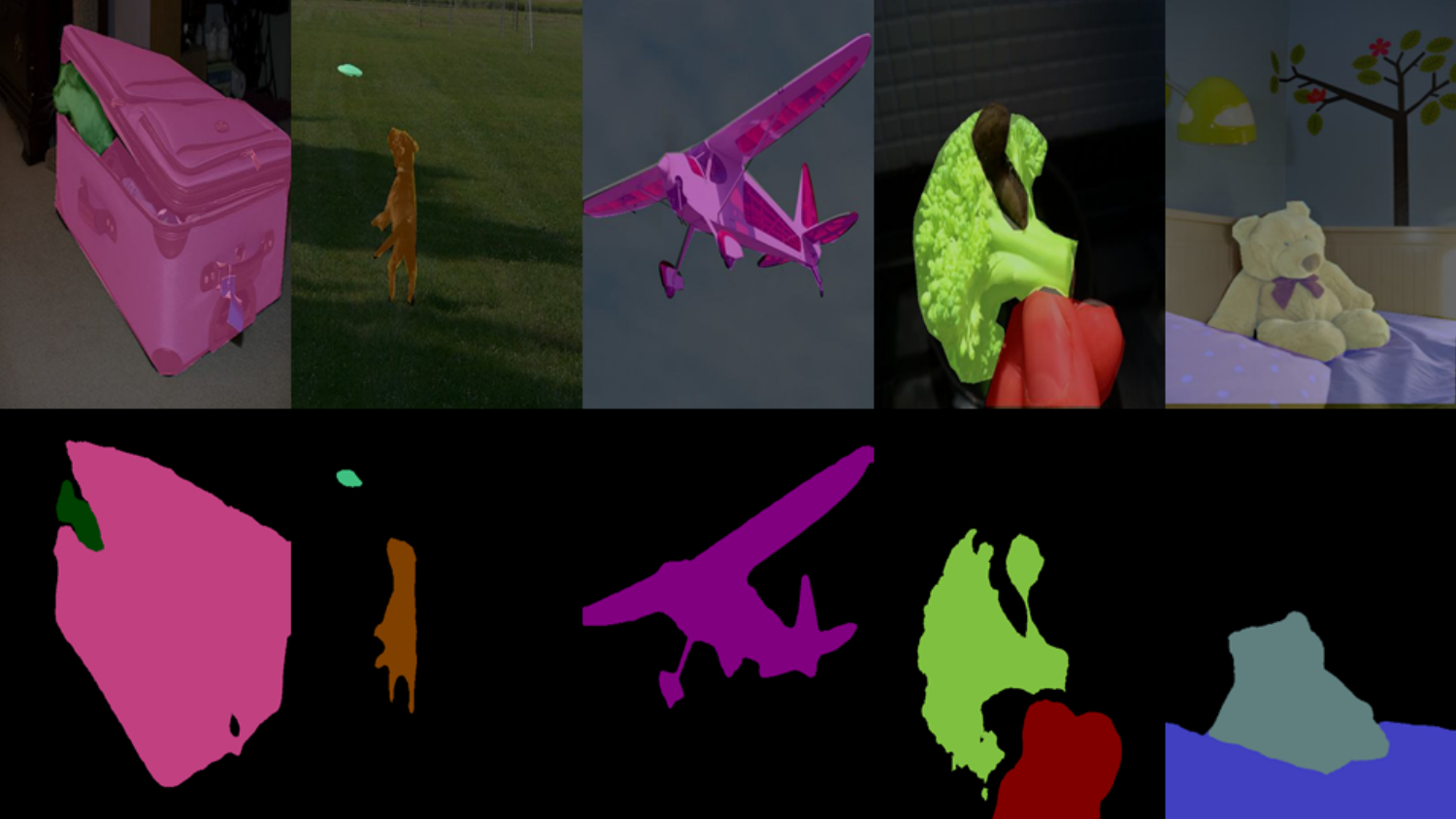}
\caption{\small Segmentation results. }
\label{fig:coco_results_2}
\end{subfigure}
\caption{\small Qualitative results of MCC on MS COCO \textit{val} set. }
\label{fig:coco_results}
\vspace{-5mm}
\end{figure}

%% file: sections/fig_seg_results.tex
\begin{figure*}[!htb]
\centering
\includegraphics[trim={0.5 0 0 0},clip,width=0.9\linewidth,height=0.45\linewidth]{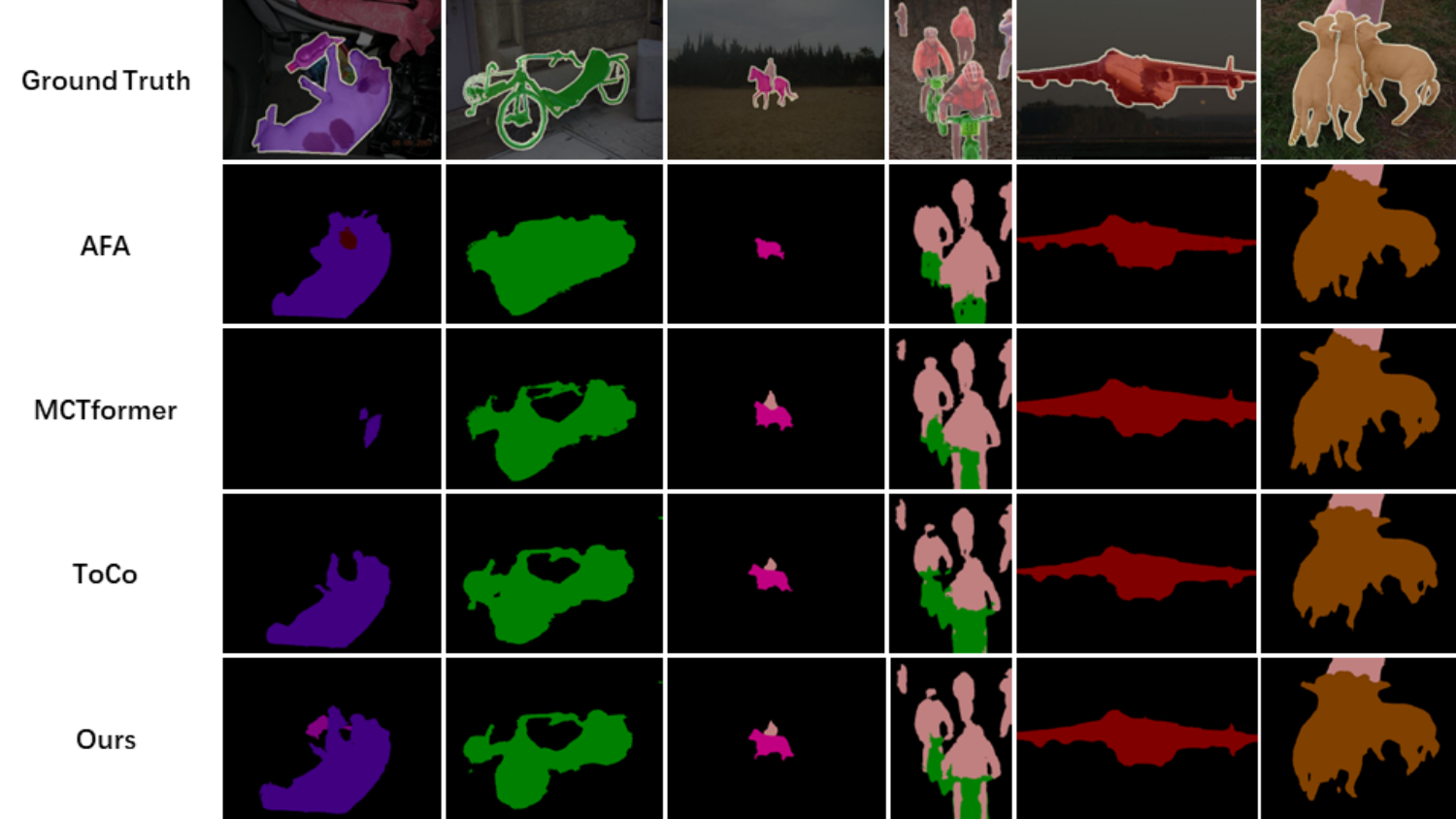}
\caption{\small Segmentation results of AFA \cite{ru2022learning}, MCTformer \cite{xu2022multi}, ToCo~\cite{ru2023token} and our method on PASCAL VOC \emph{val} set}
\label{fig:seg_results}
\vspace{-5mm}
\end{figure*}

%% file: sections/tab_ablation.tex
\begin{table}[!ht]
\caption{\small Performance comparison with baseline model in terms of mIoU(\%). MS is the commonly used multi-scale testing~\cite{CP2015Semantic} strategy and CRF is dense conditional random field~\cite{krahenbuhl2011efficient}.}
\begin{center}
\renewcommand\arraystretch{1.2}
\resizebox{1.0\linewidth}{!}{
\begin{tabular}{@{} l|p{0.4cm}<{\centering} p{1.6cm}<{\centering}| p{0.4cm}<{\centering} p{1.6cm}<{\centering}}
\hline
 \multirow{2}*{Methods} & \multicolumn{2}{c|}{VOC \emph{train}} & \multicolumn{2}{c}{VOC \emph{val}} \\
 \cline{2-5}
~  & MS & MS + CRF & MS & MS + CRF\\
\hline \hline

Baseline($\mathcal{L}_{\text{cls}} + \mathcal{L}_{\text{cls}}^{\text{aux}} + \mathcal{L}_{\text{seg}}$) & 57.0&57.2 & 55.7& 55.8 \\

Baseline + $ \mathcal{L}_{\text{aff}}$ & 69.6& 70.1 & 67.1 & 67.5\\

Baseline + $\mathcal{L}_{\text{aff}} + \mathcal{L}_{\text{mcc}}$ & \textbf{73.6} & \textbf{74.3} & \textbf{69.6} & \textbf{70.3} \\
\hline
\end{tabular}
}
\end{center}
\vspace{-5mm}
\label{table:ablation}
\end{table}


%% file: sections/fig_scale_ratio.tex
\begin{figure}[!ht]
\centering
\includegraphics[width=0.48\textwidth]{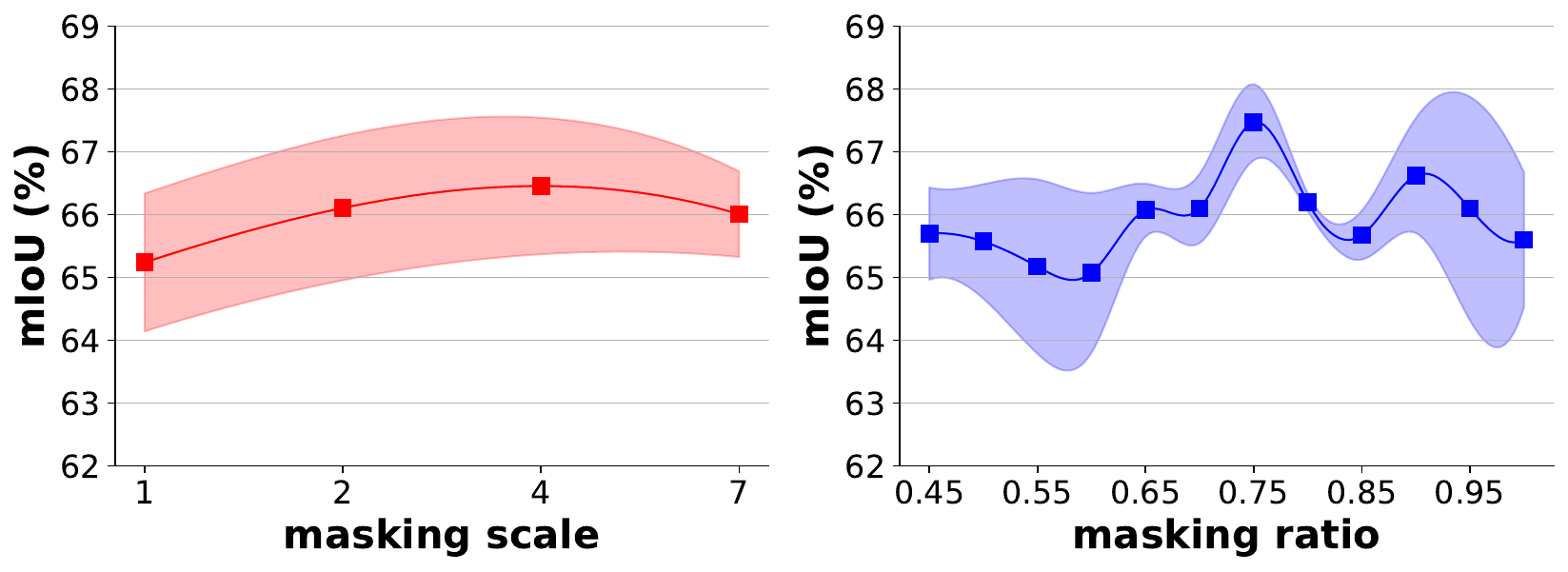}
\caption{\small Ablation Study on the impact of masking ratios on masking scales (left), and the impact of masking scales on masking ratios (right). The mean and standard deviation of the segmentation results, reported in terms of mIoU ($\%$), were obtained from the PASCAL VOC \emph{val} set \textit{without} MS inference and CRF post-processing.}
\label{fig:abl_mask}
\end{figure}

%% file: sections/tab_fg_bg_thres.tex
\begin{table}[!ht]
\caption{\small Impact of the choices of background threshold $\beta_{bg}$ and foreground threshold $\beta_{fg}$.}
\begin{center}
\renewcommand\arraystretch{1.2}
\resizebox{0.45\textwidth}{!}{
\begin{tabular}{c|p{1.6cm}<{\centering} p{1.6cm}<{\centering} p{1.6cm}<{\centering} p{1.6cm}<{\centering} p{1.6cm}<{\centering}}
\hline
 \backslashbox{$\beta_{fg}$}{$\beta_{bg}$} & 0.2 & \textbf{0.25} & 0.30 \\
\hline \hline
0.65 & 63.4 & 65.4 & 67.6 \\
\textbf{0.70} & 64.4 & \textbf{68.8} & 66.0 \\
0.75 & 67.1 & 65.7 & 64.7 \\
0.80 & 65.7 & 64.9 & 61.5 \\ 
\hline
\end{tabular}
}
\vspace{-5mm}
\end{center}
\label{table:fg_bg_thres}
\end{table}

%% file: sections/limitations.tex
\input{sections/fig_failure_cases}
\section{Limitations}

The MCC typically excels in generating high-quality CAMs. However, it does have limitations and may exhibit shortcomings in specific scenarios. Notably, we observe that it can encounter challenges in cases characterized by clutter, incompleteness, missing objects, and over-activation. These issues are visually depicted in Figure~\ref{fig:failure} for clarity. To provide a more detailed insight into these failure scenarios, it's important to highlight a few specific situations where MCC may not perform optimally. 
Understanding these limitations is crucial for effectively employing MCC in applications and scenarios where these specific failure cases might be encountered. It also underscores the need for further research and potential improvements to address these challenges.

%% file: sections/fig_failure_cases.tex
\begin{figure}[!ht]
\centering
\includegraphics[width=0.45\textwidth,height=0.2\textwidth]{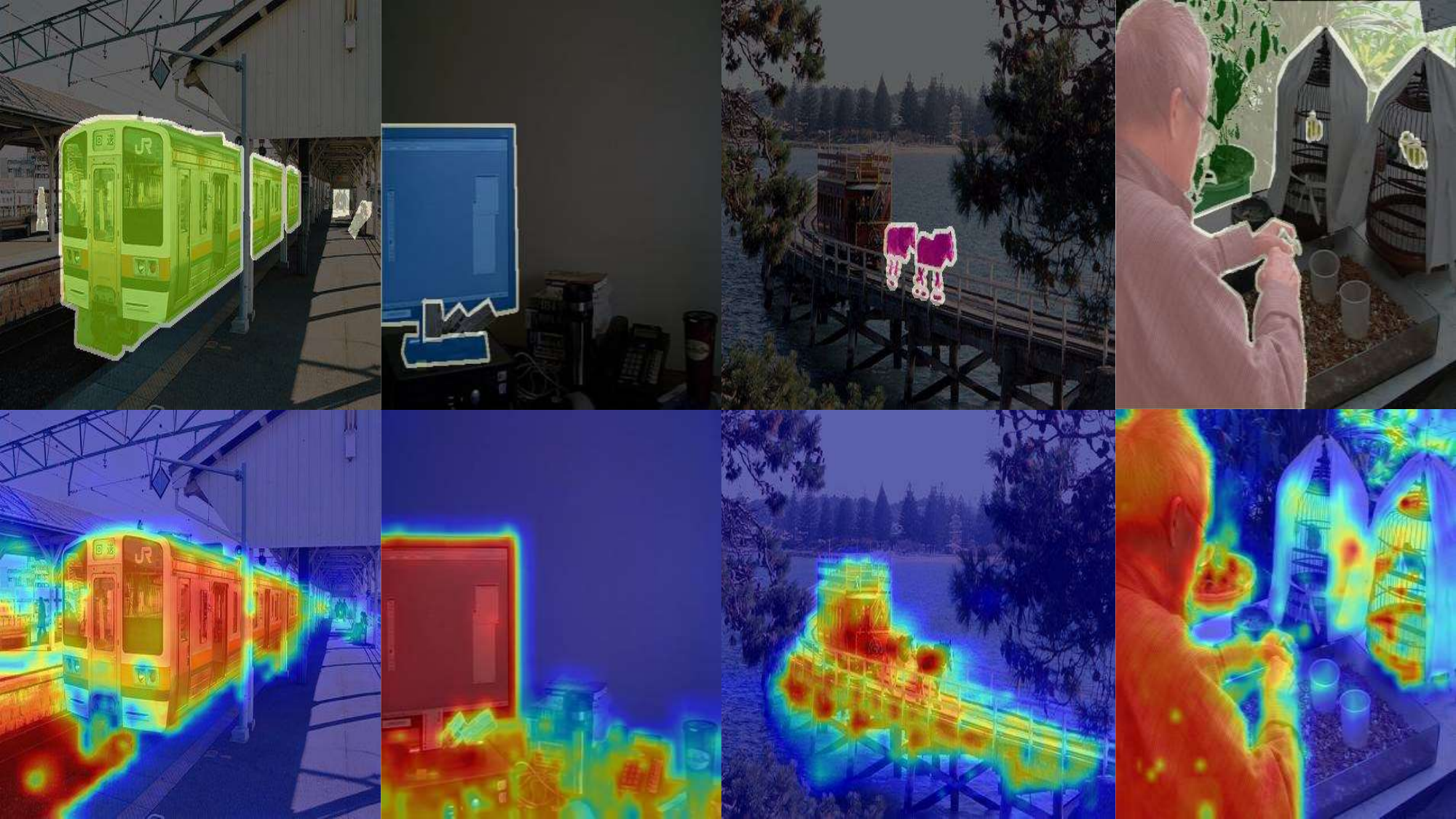}
\caption{\small Failure cases: (1) Object appearance cues are easily confused. (2) Multiple objects appear in close proximity or overlap. (3) Objects with intricate and complex structures.}
\label{fig:failure}
\end{figure}

%% file: sections/conclusion.tex
\section{Conclusion}
In this work, we draw inspiration from masked image modeling and contrastive learning to devise an effective module termed Masked Collaborative Contrast (MCC) to facilitate the performance of weakly supervised semantic segmentation. The experimental results manifest the superiority of the proposed approach over existing methods and underscore its potential for related tasks.

%% file: NewPaperCameraReady.bbl
\begin{thebibliography}{10}\itemsep=-1pt

\bibitem{ahn2019weakly}
Jiwoon Ahn, Sunghyun Cho, and Suha Kwak.
\newblock Weakly supervised learning of instance segmentation with inter-pixel
  relations.
\newblock In {\em Proceedings of the IEEE/CVF conference on computer vision and
  pattern recognition}, pages 2209--2218, 2019.

\bibitem{ahn2018learning}
Jiwoon Ahn and Suha Kwak.
\newblock Learning pixel-level semantic affinity with image-level supervision
  for weakly supervised semantic segmentation.
\newblock In {\em Proceedings of the IEEE conference on computer vision and
  pattern recognition}, pages 4981--4990, 2018.

\bibitem{araslanov2020single}
Nikita Araslanov and Stefan Roth.
\newblock Single-stage semantic segmentation from image labels.
\newblock In {\em Proceedings of the IEEE/CVF Conference on Computer Vision and
  Pattern Recognition}, pages 4253--4262, 2020.

\bibitem{bearman2016point}
Amy Bearman, Olga Russakovsky, Vittorio Ferrari, and Li Fei-Fei.
\newblock What’s the point: Semantic segmentation with point supervision.
\newblock In {\em Computer Vision--ECCV 2016: 14th European Conference,
  Amsterdam, The Netherlands, October 11--14, 2016, Proceedings, Part VII 14},
  pages 549--565. Springer, 2016.

\bibitem{caron2021emerging}
Mathilde Caron, Hugo Touvron, Ishan Misra, Herv{\'e} J{\'e}gou, Julien Mairal,
  Piotr Bojanowski, and Armand Joulin.
\newblock Emerging properties in self-supervised vision transformers.
\newblock In {\em Proceedings of the IEEE/CVF international conference on
  computer vision}, pages 9650--9660, 2021.

\bibitem{CP2015Semantic}
Liang-Chieh Chen, George Papandreou, Iasonas Kokkinos, Kevin Murphy, and Alan~L
  Yuille.
\newblock Semantic image segmentation with deep convolutional nets and fully
  connected crfs.
\newblock In {\em ICLR}, 2015.

\bibitem{chen2023extracting}
Zhaozheng Chen and Qianru Sun.
\newblock Extracting class activation maps from non-discriminative features as
  well.
\newblock In {\em Proceedings of the IEEE/CVF conference on computer vision and
  pattern recognition}, 2023.

\bibitem{chen2022class}
Zhaozheng Chen, Tan Wang, Xiongwei Wu, Xian-Sheng Hua, Hanwang Zhang, and
  Qianru Sun.
\newblock Class re-activation maps for weakly-supervised semantic segmentation.
\newblock In {\em Proceedings of the IEEE/CVF Conference on Computer Vision and
  Pattern Recognition}, pages 969--978, 2022.

\bibitem{cheng2023out}
Zesen Cheng, Pengchong Qiao, Kehan Li, Siheng Li, Pengxu Wei, Xiangyang Ji, Li
  Yuan, Chang Liu, and Jie Chen.
\newblock Out-of-candidate rectification for weakly supervised semantic
  segmentation.
\newblock In {\em Proceedings of the IEEE/CVF Conference on Computer Vision and
  Pattern Recognition}, 2023.

\bibitem{dosovitskiy2020vit}
Alexey Dosovitskiy, Lucas Beyer, Alexander Kolesnikov, Dirk Weissenborn,
  Xiaohua Zhai, Thomas Unterthiner, Mostafa Dehghani, Matthias Minderer, Georg
  Heigold, Sylvain Gelly, Jakob Uszkoreit, and Neil Houlsby.
\newblock An image is worth 16x16 words: Transformers for image recognition at
  scale.
\newblock {\em ICLR}, 2021.

\bibitem{Everingham2010ThePV}
Mark Everingham, Luc~Van Gool, Christopher K.~I. Williams, John~M. Winn, and
  Andrew Zisserman.
\newblock The pascal visual object classes (voc) challenge.
\newblock {\em International Journal of Computer Vision}, 88:303--338, 2010.

\bibitem{gao2021ts}
Wei Gao, Fang Wan, Xingjia Pan, Zhiliang Peng, Qi Tian, Zhenjun Han, Bolei
  Zhou, and Qixiang Ye.
\newblock Ts-cam: Token semantic coupled attention map for weakly supervised
  object localization.
\newblock In {\em Proceedings of the IEEE/CVF International Conference on
  Computer Vision}, pages 2886--2895, 2021.

\bibitem{6126343}
Bharath Hariharan, Pablo Arbeláez, Lubomir Bourdev, Subhransu Maji, and
  Jitendra Malik.
\newblock Semantic contours from inverse detectors.
\newblock In {\em 2011 International Conference on Computer Vision}, pages
  991--998, 2011.

\bibitem{he2023mitigating}
Jingxuan He, Lechao Cheng, Chaowei Fang, Dingwen Zhang, Zhangye Wang, and Wei
  Chen.
\newblock Mitigating undisciplined over-smoothing in transformer for weakly
  supervised semantic segmentation.
\newblock {\em arXiv preprint arXiv:2305.03112}, 2023.

\bibitem{he2020momentum}
Kaiming He, Haoqi Fan, Yuxin Wu, Saining Xie, and Ross Girshick.
\newblock Momentum contrast for unsupervised visual representation learning.
\newblock In {\em Proceedings of the IEEE/CVF conference on computer vision and
  pattern recognition}, pages 9729--9738, 2020.

\bibitem{jiang2022l2g}
Peng-Tao Jiang, Yuqi Yang, Qibin Hou, and Yunchao Wei.
\newblock L2g: A simple local-to-global knowledge transfer framework for weakly
  supervised semantic segmentation.
\newblock In {\em Proceedings of the IEEE/CVF Conference on Computer Vision and
  Pattern Recognition}, pages 16886--16896, 2022.

\bibitem{jonnarth2022imp}
Arvi Jonnarth and Michael Felsberg.
\newblock Importance sampling cams for weakly-supervised segmentation.
\newblock In {\em ICASSP 2022 - 2022 IEEE International Conference on
  Acoustics, Speech and Signal Processing (ICASSP)}, pages 2639--2643, 2022.

\bibitem{kho2022exploiting}
Sungpil Kho, Pilhyeon Lee, Wonyoung Lee, Minsong Ki, and Hyeran Byun.
\newblock Exploiting shape cues for weakly supervised semantic segmentation.
\newblock {\em Pattern Recognition}, 132:108953, 2022.

\bibitem{kolesnikov2016seed}
Alexander Kolesnikov and Christoph~H Lampert.
\newblock Seed, expand and constrain: Three principles for weakly-supervised
  image segmentation.
\newblock In {\em Computer Vision--ECCV 2016: 14th European Conference,
  Amsterdam, The Netherlands, October 11--14, 2016, Proceedings, Part IV 14},
  pages 695--711. Springer, 2016.

\bibitem{krahenbuhl2011efficient}
Philipp Kr{\"a}henb{\"u}hl and Vladlen Koltun.
\newblock Efficient inference in fully connected crfs with gaussian edge
  potentials.
\newblock {\em Advances in neural information processing systems}, 24, 2011.

\bibitem{lee2021reducing}
Jungbeom Lee, Jooyoung Choi, Jisoo Mok, and Sungroh Yoon.
\newblock Reducing information bottleneck for weakly supervised semantic
  segmentation.
\newblock {\em Advances in Neural Information Processing Systems},
  34:27408--27421, 2021.

\bibitem{lee2021anti}
Jungbeom Lee, Eunji Kim, and Sungroh Yoon.
\newblock Anti-adversarially manipulated attributions for weakly and
  semi-supervised semantic segmentation.
\newblock In {\em Proceedings of the IEEE/CVF Conference on Computer Vision and
  Pattern Recognition}, pages 4071--4080, 2021.

\bibitem{lee2021bbam}
Jungbeom Lee, Jihun Yi, Chaehun Shin, and Sungroh Yoon.
\newblock Bbam: Bounding box attribution map for weakly supervised semantic and
  instance segmentation.
\newblock In {\em Proceedings of the IEEE/CVF conference on computer vision and
  pattern recognition}, pages 2643--2652, 2021.

\bibitem{lee2022threshold}
Minhyun Lee, Dongseob Kim, and Hyunjung Shim.
\newblock Threshold matters in wsss: manipulating the activation for the robust
  and accurate segmentation model against thresholds.
\newblock In {\em Proceedings of the IEEE/CVF Conference on Computer Vision and
  Pattern Recognition}, pages 4330--4339, 2022.

\bibitem{lee2021railroad}
Seungho Lee, Minhyun Lee, Jongwuk Lee, and Hyunjung Shim.
\newblock Railroad is not a train: Saliency as pseudo-pixel supervision for
  weakly supervised semantic segmentation.
\newblock In {\em Proceedings of the IEEE/CVF conference on computer vision and
  pattern recognition}, pages 5495--5505, 2021.

\bibitem{li2023boosting}
Hao Li, Dingwen Zhang, Nian Liu, Lechao Cheng, Yalun Dai, Chao Zhang, Xinggang
  Wang, and Junwei Han.
\newblock Boosting low-data instance segmentation by unsupervised pre-training
  with saliency prompt.
\newblock In {\em Proceedings of the IEEE/CVF Conference on Computer Vision and
  Pattern Recognition}, pages 15485--15494, 2023.

\bibitem{liexpansion}
JINLONG LI, ZEQUN JIE, Xu Wang, Lin Ma, et~al.
\newblock Expansion and shrinkage of localization for weakly-supervised
  semantic segmentation.
\newblock In {\em Advances in Neural Information Processing Systems}, 2022.

\bibitem{li2023transcam}
Ruiwen Li, Zheda Mai, Zhibo Zhang, Jongseong Jang, and Scott Sanner.
\newblock Transcam: Transformer attention-based cam refinement for weakly
  supervised semantic segmentation.
\newblock {\em Journal of Visual Communication and Image Representation},
  92:103800, 2023.

\bibitem{lin2014microsoft}
Tsung-Yi Lin, Michael Maire, Serge Belongie, James Hays, Pietro Perona, Deva
  Ramanan, Piotr Doll{\'a}r, and C~Lawrence Zitnick.
\newblock Microsoft coco: Common objects in context.
\newblock In {\em Computer Vision--ECCV 2014: 13th European Conference, Zurich,
  Switzerland, September 6-12, 2014, Proceedings, Part V 13}, pages 740--755.
  Springer, 2014.

\bibitem{lin2022clip}
Yuqi Lin, Minghao Chen, Wenxiao Wang, Boxi Wu, Ke Li, Binbin Lin, Haifeng Liu,
  and Xiaofei He.
\newblock Clip is also an efficient segmenter: A text-driven approach for
  weakly supervised semantic segmentation.
\newblock {\em arXiv preprint arXiv:2212.09506}, 2022.

\bibitem{loshchilov2018decoupled}
Ilya Loshchilov and Frank Hutter.
\newblock Decoupled weight decay regularization.
\newblock In {\em International Conference on Learning Representations}, 2019.

\bibitem{oord2018representation}
Aaron van~den Oord, Yazhe Li, and Oriol Vinyals.
\newblock Representation learning with contrastive predictive coding.
\newblock {\em arXiv preprint arXiv:1807.03748}, 2018.

\bibitem{pan2022learning}
Junwen Pan, Pengfei Zhu, Kaihua Zhang, Bing Cao, Yu Wang, Dingwen Zhang, Junwei
  Han, and Qinghua Hu.
\newblock Learning self-supervised low-rank network for single-stage weakly and
  semi-supervised semantic segmentation.
\newblock {\em International Journal of Computer Vision}, 130(5):1181--1195,
  2022.

\bibitem{peng2021conformer}
Zhiliang Peng, Wei Huang, Shanzhi Gu, Lingxi Xie, Yaowei Wang, Jianbin Jiao,
  and Qixiang Ye.
\newblock Conformer: Local features coupling global representations for visual
  recognition.
\newblock In {\em Proceedings of the IEEE/CVF international conference on
  computer vision}, pages 367--376, 2021.

\bibitem{Qin_Wu_Xiao_Li_Wang_2022}
Jie Qin, Jie Wu, Xuefeng Xiao, Lujun Li, and Xingang Wang.
\newblock Activation modulation and recalibration scheme for weakly supervised
  semantic segmentation.
\newblock {\em Proceedings of the AAAI Conference on Artificial Intelligence},
  36(2):2117--2125, Jun. 2022.

\bibitem{rossetti2022max}
Simone Rossetti, Damiano Zappia, Marta Sanzari, Marco Schaerf, and Fiora Pirri.
\newblock Max pooling with vision transformers reconciles class and shape in
  weakly supervised semantic segmentation.
\newblock In {\em Computer Vision--ECCV 2022: 17th European Conference, Tel
  Aviv, Israel, October 23--27, 2022, Proceedings, Part XXX}, pages 446--463.
  Springer, 2022.

\bibitem{ru2022learning}
Lixiang Ru, Yibing Zhan, Baosheng Yu, and Bo Du.
\newblock Learning affinity from attention: end-to-end weakly-supervised
  semantic segmentation with transformers.
\newblock In {\em Proceedings of the IEEE/CVF Conference on Computer Vision and
  Pattern Recognition}, pages 16846--16855, 2022.

\bibitem{ru2023token}
Lixiang Ru, Heliang Zheng, Yibing Zhan, and Bo Du.
\newblock Token contrast for weakly-supervised semantic segmentation.
\newblock In {\em CVPR}, 2023.

\bibitem{russakovsky2015imagenet}
Olga Russakovsky, Jia Deng, Hao Su, Jonathan Krause, Sanjeev Satheesh, Sean Ma,
  Zhiheng Huang, Andrej Karpathy, Aditya Khosla, Michael Bernstein, et~al.
\newblock Imagenet large scale visual recognition challenge.
\newblock {\em International journal of computer vision}, 115:211--252, 2015.

\bibitem{shi2022revisiting}
Han Shi, Jiahui Gao, Hang Xu, Xiaodan Liang, Zhenguo Li, Lingpeng Kong, Stephen
  Lee, and James~T Kwok.
\newblock Revisiting over-smoothing in bert from the perspective of graph.
\newblock {\em arXiv preprint arXiv:2202.08625}, 2022.

\bibitem{su2022re}
Hui Su, Yue Ye, Zhiwei Chen, Mingli Song, and Lechao Cheng.
\newblock Re-attention transformer for weakly supervised object localization.
\newblock In {\em The 33rd British Machine Vision Conference (BMVC), 2022},
  2022.

\bibitem{su2022sasformer}
Hui Su, Yue Ye, Wei Hua, Lechao Cheng, and Mingli Song.
\newblock Sasformer: Transformers for sparsely annotated semantic segmentation.
\newblock In {\em IEEE International Conference on Multimedia and Expo (ICME),
  2023}, 2022.

\bibitem{touvron2021training}
Hugo Touvron, Matthieu Cord, Matthijs Douze, Francisco Massa, Alexandre
  Sablayrolles, and Herv{\'e} J{\'e}gou.
\newblock Training data-efficient image transformers \& distillation through
  attention.
\newblock In {\em International conference on machine learning}, pages
  10347--10357. PMLR, 2021.

\bibitem{wang2020self}
Yude Wang, Jie Zhang, Meina Kan, Shiguang Shan, and Xilin Chen.
\newblock Self-supervised equivariant attention mechanism for weakly supervised
  semantic segmentation.
\newblock In {\em Proceedings of the IEEE/CVF Conference on Computer Vision and
  Pattern Recognition}, pages 12275--12284, 2020.

\bibitem{wu2021embed}
Tong Wu, Junshi Huang, Guangyu Gao, Xiaoming Wei, Xiaolin Wei, Xuan Luo, and
  Chi~Harold Liu.
\newblock Embedded discriminative attention mechanism for weakly supervised
  semantic segmentation.
\newblock In {\em Proceedings of the IEEE/CVF Conference on Computer Vision and
  Pattern Recognition (CVPR)}, pages 16765--16774, June 2021.

\bibitem{xie2022clims}
Jinheng Xie, Xianxu Hou, Kai Ye, and Linlin Shen.
\newblock Clims: Cross language image matching for weakly supervised semantic
  segmentation.
\newblock In {\em Proceedings of the IEEE/CVF Conference on Computer Vision and
  Pattern Recognition (CVPR)}, pages 4483--4492, June 2022.

\bibitem{xie2022simmim}
Zhenda Xie, Zheng Zhang, Yue Cao, Yutong Lin, Jianmin Bao, Zhuliang Yao, Qi
  Dai, and Han Hu.
\newblock Simmim: A simple framework for masked image modeling.
\newblock In {\em Proceedings of the IEEE/CVF Conference on Computer Vision and
  Pattern Recognition}, pages 9653--9663, 2022.

\bibitem{xu2021leveraging}
Lian Xu, Wanli Ouyang, Mohammed Bennamoun, Farid Boussaid, Ferdous Sohel, and
  Dan Xu.
\newblock Leveraging auxiliary tasks with affinity learning for weakly
  supervised semantic segmentation.
\newblock In {\em Proceedings of the IEEE/CVF International Conference on
  Computer Vision}, pages 6984--6993, 2021.

\bibitem{xu2022multi}
Lian Xu, Wanli Ouyang, Mohammed Bennamoun, Farid Boussaid, and Dan Xu.
\newblock Multi-class token transformer for weakly supervised semantic
  segmentation.
\newblock In {\em Proceedings of the IEEE/CVF Conference on Computer Vision and
  Pattern Recognition}, pages 4310--4319, 2022.

\bibitem{xue2022protopformer}
Mengqi Xue, Qihan Huang, Haofei Zhang, Lechao Cheng, Jie Song, Minghui Wu, and
  Mingli Song.
\newblock Protopformer: Concentrating on prototypical parts in vision
  transformers for interpretable image recognition.
\newblock {\em arXiv preprint arXiv:2208.10431}, 2022.

\bibitem{yao2021non}
Yazhou Yao, Tao Chen, Guo-Sen Xie, Chuanyi Zhang, Fumin Shen, Qi Wu, Zhenmin
  Tang, and Jian Zhang.
\newblock Non-salient region object mining for weakly supervised semantic
  segmentation.
\newblock In {\em Proceedings of the IEEE/CVF Conference on Computer Vision and
  Pattern Recognition}, pages 2623--2632, 2021.

\bibitem{zhang2020reliability}
Bingfeng Zhang, Jimin Xiao, Yunchao Wei, Mingjie Sun, and Kaizhu Huang.
\newblock Reliability does matter: An end-to-end weakly supervised semantic
  segmentation approach.
\newblock In {\em Proceedings of the AAAI Conference on Artificial
  Intelligence}, volume~34, pages 12765--12772, 2020.

\bibitem{zhang2021dynamic}
Bingfeng Zhang, Jimin Xiao, and Yao Zhao.
\newblock Dynamic feature regularized loss for weakly supervised semantic
  segmentation.
\newblock {\em arXiv preprint arXiv:2108.01296}, 2021.

\bibitem{zhang2021weakly}
Dingwen Zhang, Wenyuan Zeng, Guangyu Guo, Chaowei Fang, Lechao Cheng, Ming-Ming
  Cheng, and Junwei Han.
\newblock Weakly supervised semantic segmentation via alternative self-dual
  teaching.
\newblock {\em arXiv preprint arXiv:2112.09459}, 2021.

\bibitem{zhou2016learning}
Bolei Zhou, Aditya Khosla, Agata Lapedriza, Aude Oliva, and Antonio Torralba.
\newblock Learning deep features for discriminative localization.
\newblock In {\em Proceedings of the IEEE conference on computer vision and
  pattern recognition}, pages 2921--2929, 2016.

\bibitem{zhu2023weaktr}
Lianghui Zhu, Yingyue Li, Jieming Fang, Yan Liu, Hao Xin, Wenyu Liu, and
  Xinggang Wang.
\newblock Weaktr: Exploring plain vision transformer for weakly-supervised
  semantic segmentation.
\newblock {\em arXiv preprint arXiv:2304.01184}, 2023.

\end{thebibliography}
